%% file: neurips_2026.tex
\documentclass{article}

\PassOptionsToPackage{numbers, compress}{natbib}
 \usepackage[preprint]{neurips_2026}

\usepackage[utf8]{inputenc} 
\usepackage[T1]{fontenc}    
\usepackage{hyperref}       
\usepackage{url}            
\usepackage{booktabs}       
\usepackage{amsfonts}       
\usepackage{nicefrac}       
\usepackage{microtype}      
\usepackage{xcolor}         
\usepackage{graphicx}
\usepackage{subcaption}
\usepackage{multirow}
\usepackage{colortbl}
\usepackage{array}
\usepackage{amsmath}
\usepackage{amssymb}
\usepackage{enumitem}
\usepackage{cleveref}
\usepackage{wrapfig}
\usepackage{graphicx}

\definecolor{myteal}{HTML}{15797A}

\title{Mix-Quant: Quantized Prefilling, Precise Decoding  \\for Agentic LLMs }

\author{%
  Haiquan Lu, \ Zigeng Chen, \ Gongfan Fang, \ Xinyin Ma,
\ Xinchao Wang\thanks{Corresponding Author} \\
  National University of Singapore\\
  \texttt{haiquanlu@u.nus.edu, xinchao@nus.edu.sg} \\
}

\begin{document}

\maketitle

\input{sections/abstract}
\input{sections/introduction}
\input{sections/related_work}
\input{sections/method}
\input{sections/experiments}
\input{sections/conclusion}

\bibliographystyle{plainnat}
\bibliography{references}

\newpage
\appendix




\end{document}

%% file: sections/abstract.tex
\begin{abstract}
LLM agents have recently emerged as a powerful paradigm for solving complex tasks through planning, tool use, memory retrieval, and multi-step interaction. However, these agentic workflows often introduce substantial input-side overhead, making the compute-intensive prefilling stage a key bottleneck in long-context, multi-turn inference. 
In this work, we propose Mix-Quant, a simple and effective phase-aware quantization framework for fast agentic inference. We first investigate FP4 quantization in agentic LLM workflows and observe that quantizing the entire inference process can incur significant performance degradation. In contrast, the prefilling stage exhibits substantial quantization redundancy and can therefore be quantized with minimal accuracy loss, despite being the dominant source of computation. Based on this insight, we apply high-throughput NVFP4 quantization to the prefilling phase while preserving BF16 precision for decoding. By decoupling prefilling acceleration from decoding quality, Mix-Quant combines phase-aware algorithmic quantization with hardware-efficient NVFP4 execution to alleviate the inference bottleneck in LLM agents. 
Extensive experiments across long-context and agentic benchmarks demonstrate that Mix-Quant largely preserves task performance while delivering significant efficiency improvements, achieving up to a 3$\times$ speedup during prefilling. Code is available at: \textcolor{magenta}{\url{https://github.com/haiquanlu/Mix-Quant}}
\end{abstract}

%% file: sections/introduction.tex
\section{Introduction}

Large language model (LLM) agents have emerged as a powerful paradigm for solving complex real-world tasks involving tool use, memory retrieval, code generation, and multi-step interaction~\citep{yao2022react, schick2023toolformer, yang2024swe, xu2025mem}. They have shown strong potential across coding agents, personal assistants, web agents, and general-purpose autonomous systems~\cite{luo2025large}. However, agentic workflows typically require repeated inference steps and multi-call interaction loops, leading to substantial context-processing overhead. In many cases, the input context can be tens to hundreds of times longer than the generated output, making the compute-intensive prefilling phase a major efficiency bottleneck in terms of both latency and throughput.

\begin{figure}
    \centering
    \includegraphics[width=1\linewidth]{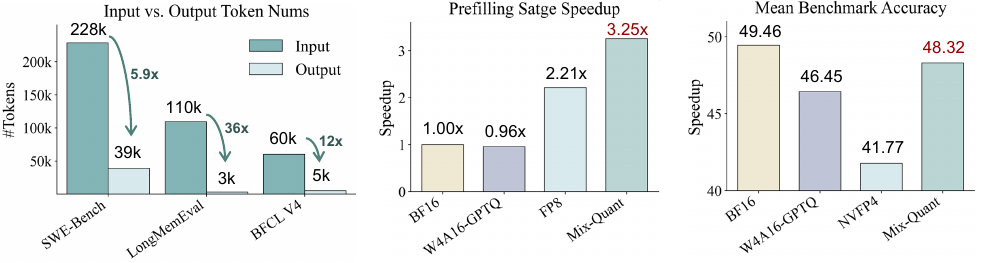}
    \caption{Agentic workflows are highly input-heavy, introducing substantial prefilling overhead. NVFP4 quantization can greatly accelerate computation, but applying it to both prefilling and decoding causes notable accuracy degradation. Mix-Quant instead uses NVFP4 for prefilling and precise BF16 for decoding, achieving substantial speedup while largely preserving agentic performance.}
    \label{fig:motivations}
\end{figure}

To alleviate the inference overhead, prior work has explored various model-efficiency strategies~\citep{frantar2022gptq, ma2023llm, lu2025mixreasoning, gu2024minillm}. While these methods are promising for improving deployment efficiency, different strategies address distinct inference bottlenecks, and applying aggressive compression uniformly across inference phases can lead to non-negligible performance degradation.
For example, post-training quantization (PTQ) is widely used due to its practicality. Weight-only PTQ~\citep{frantar2022gptq, lin2024awq} lowers memory footprint by representing model weights in low-bit formats such as INT4, improving throughput in memory-bound autoregressive decoding. However, it provides limited acceleration for the compute-bound prefill phase because activations remain in high precision. In contrast, weight-and-activation quantization enables~\cite{xiao2023smoothquant} low-bit matrix multiplications, directly reducing computational cost, but it can degrade performance, especially on complex long-trajectory tasks, because errors accumulate at each decoding step~\citep{li2025quantization, zhao2025qspec}.

Applying a uniform quantization strategy to both prefilling and decoding often leads to an unfavorable efficiency-performance trade-off. Prefilling and decoding serve different roles in LLM inference and exhibit distinct efficiency bottlenecks~\cite{zhong2024distserve, patel2024splitwise}. More specifically, prefilling processes a fixed input context and is compute-intensive, while decoding generates tokens autoregressively and is more sensitive to accumulated numerical errors. This distinction raises an important question: \textit{Can we decouple prefilling and decoding, and optimize the model for each phase according to its distinct characteristics?}

Motivated by this, we propose Mix-Quant, a phase-aware quantization framework for efficient long-context LLM agentic inference. Mix-Quant applies high-throughput NVFP4 weight-and-activation quantization to the compute-intensive prefilling phase, while preserving BF16 precision for autoregressive decoding. 
NVFP4 precision~\citep{nvidia2025nvfp4} is a microscaling FP4 format introduced with NVIDIA Blackwell, which uses fine-grained scaling to improve numerical accuracy at ultra-low bit-widths and provides native hardware support for efficient low-precision computation. 
The design of our method rests on two key observations: (1) Long-context, multi-turn agentic workflows introduce substantial context-processing overhead, making the compute-intensive prefilling phase a major efficiency bottleneck. Therefore, optimizing the prefilling phase is critical for efficient agentic inference; (2) Prefilling and decoding exhibit distinct computational bottlenecks and quantization redundancy behaviors. Prefilling processes a fixed input sequence in parallel and is suited to aggressive quantization: quantization errors do not recursively affect future inputs within the same prefill pass, and long agentic contexts often contain substantial redundancy. In contrast, decoding is much more error-sensitive, as each sampled token affects the generation process. Quantization errors can thus propagate and accumulate over long trajectories, ultimately degrading final task performance.
By integrating high-throughput NVFP4 computation into prefilling while keeping decoding in precise BF16, Mix-Quant combines algorithm-level phase awareness with hardware-level acceleration, addressing the long-context processing bottleneck in agentic inference while preserving overall task performance.

We evaluate Mix-Quant on a comprehensive suite of long-context and agentic benchmarks, including two widely used long-context benchmarks~\citep{bai2024longbenchv2,artificialanalysis2025lcr} and three multi-turn agentic benchmarks~\citep{ patil2025berkeley, wu2025longmemeval, barres2025tau}, with state-of-the-art agentic base models~\citep{gemma4_model_card, qwen3.5, qwen3}. The results show that Mix-Quant can largely preserve task performance across diverse long-context and agentic scenarios compared with uniform NVFP4 quantization, while achieving a 2--3$\times$ prefill speedup across varying sequence lengths and batch sizes. These findings demonstrate that phase-aware quantization provides a favorable efficiency-performance trade-off for input-heavy LLM agentic inference.

\textbf{Contributions.} To summarize, our main contributions include: (1) We reveal that LLM agentic workflows are highly input-heavy due to multi-step interactions with environments, making the compute-intensive prefilling phase a major efficiency bottleneck in long-context agentic inference. Meanwhile, naive model-efficiency methods can hurt task performance, highlighting the need for phase-aware optimization. (2) We propose Mix-Quant, a phase-aware quantization framework that applies NVFP4 quantization to prefilling while retaining BF16 precision for autoregressive decoding, thereby improving efficiency without introducing severe error accumulation during generation. (3) We empirically show that Mix-Quant largely preserves agentic task performance while significantly improving inference efficiency, achieving up to a 3$\times$ prefill speedup over BF16 and demonstrating the potential of phase-aware model quantization for efficient and reliable LLM agents.

%% file: sections/related_work.tex
\section{Related Work}
\textbf{Long-Context Agentic Workflows.} LLM agents extend language models with action interfaces, external tools, memory, and feedback from interactive environments. ReAct introduced the pattern of interleaving reasoning traces with environment actions \citep{yao2023react}, while Toolformer showed that language models can learn to invoke external APIs and condition on tool outputs \citep{schick2023toolformer}. WebGPT demonstrated browser-assisted question answering \citep{nakano2021webgpt}, MemGPT explored memory management for long-lived interactions \citep{packer2023memgpt}, and SWE-agents~\citep{jimenez2023swe, yang2024sweagent} showed the importance of agent-computer interfaces for software engineering. These systems repeatedly call an LLM with prompts that include instructions, tool schemas, retrieved evidence, execution traces, and memory states. Recent work on agentic inference further emphasize substantial input-token overhead, repeated-context redundancy, and high serving cost \citep{wadlom2026efficient, wu2026plena}. Mix-Quant is motivated by this workload shift: for long-context agents, accelerating context processing is often as important as improving token generation throughput.

\textbf{Prefill-Decode Disaggregation.}
Prefill and decode have different computational profiles. Prefill processes all prompt tokens in parallel and is dominated by large matrix multiplications, whereas decode advances autoregressively and repeatedly streams model weights and KV-cache entries. Serving systems have exploited this distinction by separating prompt processing from token generation. Splitwise maps prefill and decode to different machine configurations \citep{patel2023splitwise}; DistServe disaggregates the phases across GPU pools to reduce interference between time-to-first-token and time-per-output-token objectives \citep{zhong2024distserve}; and other algorithmic approaches optimize long-context prompt processing through model transformations~\citep{qiao2025swiftkv} and dynamic sparse attention~\citep{jiang2024minference, fan2026flashprefill} for faster prefill. These works show that prefill should be treated as a distinct system-level workload. Mix-Quant is naturally compatible with prefill-decode disaggregated serving: the quantized prefill path can be deployed on prefill workers, while the high-precision decoding path remains on decode workers. 
Moreover, Mix-Quant is complementary to sparse-attention optimization methods and can be combined with them to further reduce long-context prefilling cost.

\textbf{Quantization for LLM Inference.} Quantization is widely used to reduce LLM inference cost \citep{zhou2024survey}. Weight-only methods such as GPTQ~\citep{frantar2023gptq} and AWQ~\citep{lin2024awq} lower memory traffic and are effective for bandwidth-bound decoding, but provide limited speedup for long-context prefill because activations remain high precision and computation is not fully executed in low bit-widths. Weight-and-activation quantization can accelerate compute-bound prefilling, but applying aggressive W4A4 quantization to the full autoregressive process is brittle, as activation errors may perturb token choices and accumulate over generation~\citep{dettmers2022llmint8, xiao2023smoothquant, zhao2024atom}. Mix-Quant therefore quantizes only context encoding while keeping decoding on the original high-precision path. To do so efficiently, it leverages NVFP4, a Blackwell-supported microscaling FP4 format that improves 4-bit fidelity through fine-grained local scaling and native hardware execution. Following recent observations that scale treatment is critical for FP4 quality \citep{egiazarian2025bridginggappromiseperformance}, Mix-Quant uses a simple hardware-aligned W4A4 prefill path with scale optimization.

%% file: sections/method.tex
\section{Method}
\label{sec:method}

\begin{figure}
    \centering
    \includegraphics[width=0.95\linewidth]{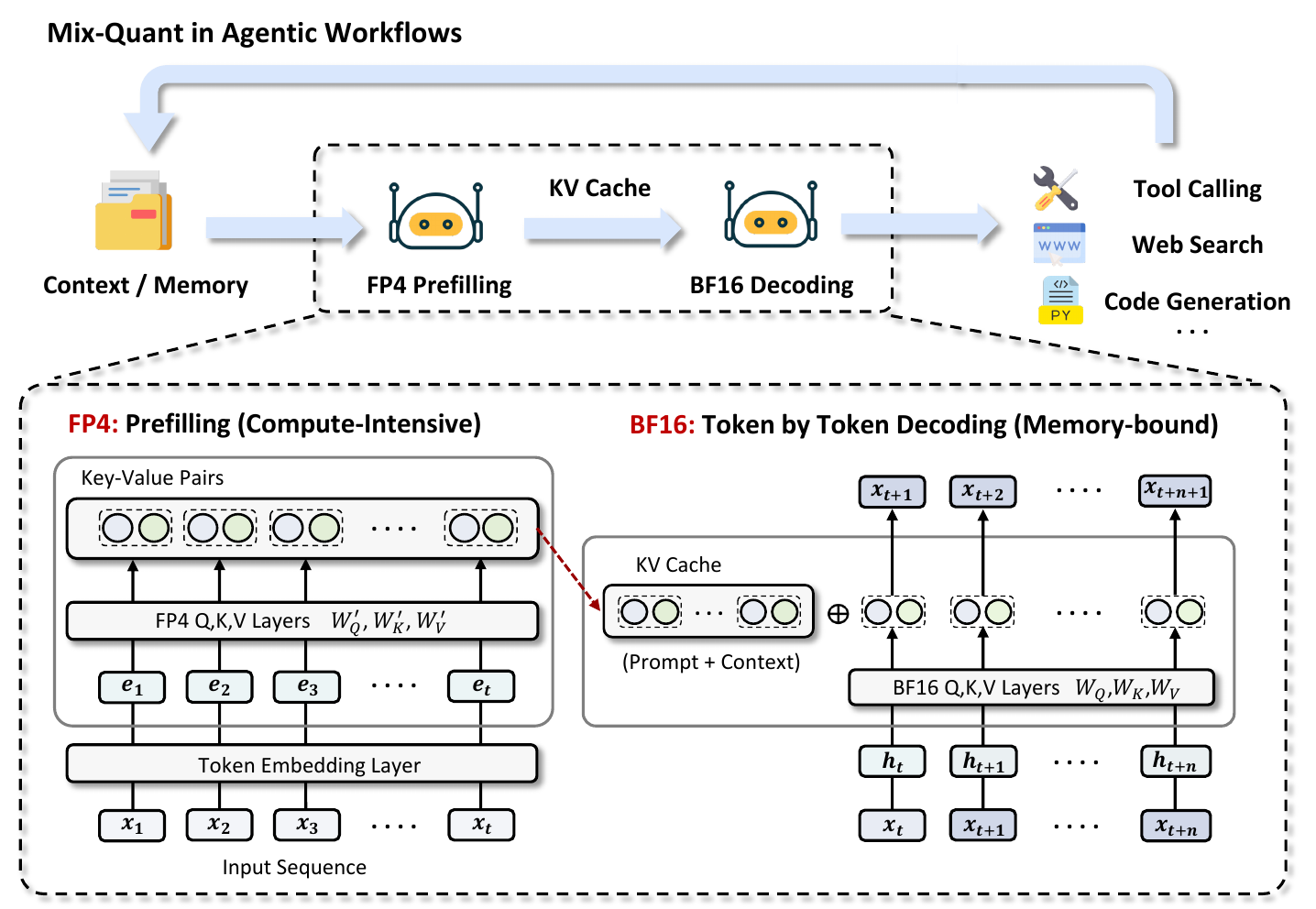}
    \caption{\textbf{Overview of Mix-Quant for efficient agentic LLM inference. }Agentic workflows repeatedly incorporate tool outputs, memory retrievals, and intermediate results into the input context, making the prefilling stage increasingly compute-intensive. Mix-Quant adopts a phase-aware quantization strategy: the context prefilling phase is accelerated with high-throughput NVFP4 computation, while autoregressive token-by-token decoding remains in BF16 to avoid error accumulation and preserve generation quality. }
    \label{fig:medthod}
\end{figure}

LLM agentic workflows can effectively solve various complex real-world tasks through multi-round interactions with external environments, tools, and memory. However, such interaction-intensive workflows substantially increase the input context that must be processed at each inference step, leading to heavy inference costs. A naive application of model-efficiency techniques to accelerate inference often compromises overall task quality and destabilizes the generation process, especially in long agentic trajectories. To address this dilemma, we introduce a decoupled model-efficiency framework that applies FP4 quantization exclusively to high-throughput context prefilling while preserving high-precision decoding for stable and effective agentic generation. 

In the remainder of this section, we first characterize the behaviour of long-context agentic workflows and identify their key efficiency bottlenecks. Next, we investigate FP4-quantized inference for both prefilling and decoding, with a particular focus on the error accumulation risks for long agentic trajectories. Finally, we detail our phase-aware quantization framework, which enables efficient and effective long-context agentic inference.

\subsection{Efficiency Bottlenecks in Long-Context Agentic Workflows.}

LLM-based agentic workflows typically solve a task through multiple rounds of model calls, tool invocations, environment observations, and memory updates. At each round, the model input may include the original user instruction, system prompt, tool descriptions, retrieved documents, previous actions, execution results, and intermediate reasoning states. As the interaction proceeds, these components are repeatedly carried over and appended to the prompt, causing the input context to grow rapidly. As shown in~\cref{fig:motivations}, the number of input tokens can be tens of times larger than that of generated output tokens.

This input-heavy characteristic makes agentic inference fundamentally different from standard single-turn generation. In conventional generation workloads, the dominant cost often comes from decoding a long output sequence. In contrast, agentic workflows usually generate only a small number of tokens at each step, such as a tool call, a short reasoning segment, or an action command, while repeatedly processing a much longer context. As a result, the overall inference cost is dominated not only by autoregressive decoding, but also, and often more critically, by repeated context prefilling.

The distinction between prefilling and decoding is illustrated in~\cref{fig:medthod}. During prefilling, the model encodes a long fixed input context and constructs the corresponding key-value cache. This stage is highly parallelizable but involves large-scale matrix multiplications across the entire context, making it compute-intensive and placing substantial pressure on accelerator compute resources. By contrast, decoding generates new tokens autoregressively, typically one token at a time. Its efficiency is often constrained by memory access and key-value cache I/O rather than by dense computation alone.

This phase-level difference also explains why many existing LLM quantization methods are insufficient for long-context agentic workflows. Prior weight-only quantization approaches~\citep{frantar2022gptq,lin2024awq} primarily reduce model weight storage and memory bandwidth, thereby improving decoding efficiency. However, because prefilling remains dominated by dense matrix computation over long contexts and dequantization overhead, weight-only quantization provides limited acceleration for the prefill stage as illustrated in~\cref{fig:motivations}. Consequently, these methods are less effective when the main bottleneck comes from repeatedly processing long input contexts, as in agentic workflows.

Rather than applying a single model-efficiency strategy uniformly to both prefilling and decoding, an effective system should adopt a phase-aware design that tailors optimization strategies to the distinct computational characteristics, task requirements, and efficiency bottlenecks of each inference stage. In this work, we take a first step toward phase-aware model efficiency by studying quantization for long-context agentic inference.

\subsection{Error Accumulation Risks of Quantized Generation.}

Model quantization is attractive for accelerating LLM inference because it can reduce memory usage and enable low-bit computation. In particular, applying weight and activation FP4 quantization to prefilling can reduce the cost of processing long input contexts, since the prefill phase is dominated by large matrix multiplications. However, naively applying FP4 quantization to the entire inference pipeline, including autoregressive decoding, can introduce significant quality degradation.

The key issue is that prefilling and decoding propagate quantization errors in different ways. During prefilling, the input context is fixed. Quantization errors may affect the hidden states and the constructed KV cache, but they do not change the input tokens being processed. Therefore, the error introduced in prefilling is mainly a representation-level perturbation on a fixed context.

\begin{wrapfigure}{r}{0.35\linewidth}
    \centering
    \includegraphics[width=1.0\linewidth]{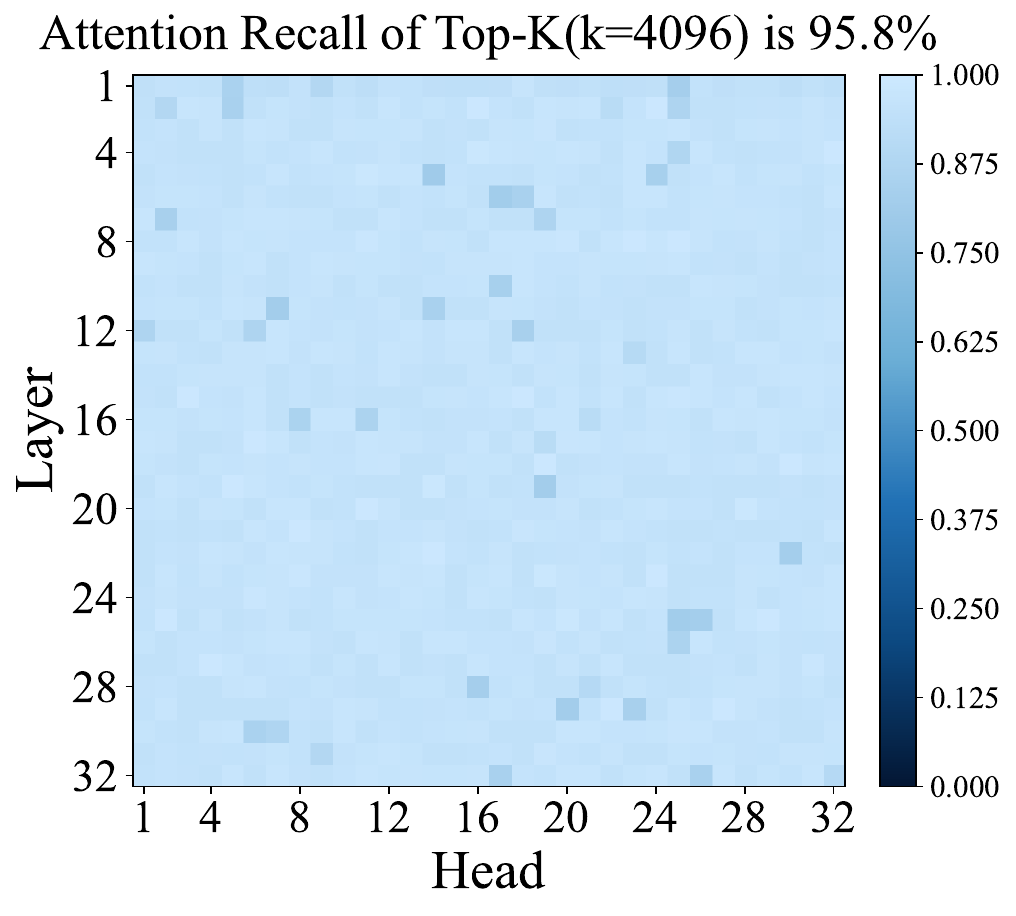}
    \caption{Attention mass concentration in a 128K-token context. The top 4,096 tokens, representing only 3.125\% of the full 128K-token context, account for an average of 95.8\% of the total attention mass. This suggests that long-context attention is highly concentrated on a small subset of tokens.}
    \label{fig:attention_mass}
\end{wrapfigure}

Moreover, long-context inputs often contain substantial redundancy~\citep{jiang2024longllmlingua}. As shown in~\cref{fig:attention_mass}, only a small set of heavy-hitter tokens dominates the attention mass at each decoding step. In the 128K-context setting, the top-4096 tokens, corresponding to only $3.125\%$ of the full context, retain $95.8\%$ of the total attention mass on average across layers and heads. This suggests that subsequent decoding is mainly influenced by a small subset of context tokens, while most tokens receive negligible attention and have limited impact on the next-token representation.
The attention mass concentration further implies that prefill-stage KV errors are not simply accumulated over all context tokens. Since the attention output is a normalized weighted aggregation over cached values, quantization errors on low-attention tokens are attenuated by their small attention weights. 
Therefore, prefill quantization errors do not simply grow linearly or explosively with the context prefilling length, which helps explain the robustness of aggressive quantization during prefilling.

In contrast, decoding is a sequential decision process. At each step, the model predicts the next token based on all previous tokens:
\begin{equation}
    y_t \sim p(y_t \mid x_{1:L}, y_{<t}).
\end{equation}
When decoding is performed under a quantized model, numerical perturbations can change the output distribution.
Even a small change in the token distribution may lead to a different sampled or selected token. Once this happens, the generated sequence diverges from the high-precision trajectory, and all future predictions are conditioned on a different history. As a result, decoding errors can accumulate over time rather than remaining local. Previous work~\citep{zhao2025qspec, li2025quantization} also observes that token prediction changes can trigger a snowball effect.

This risk is amplified in long agentic trajectories. A single erroneous token may produce an invalid tool call, select a wrong action, corrupt a code edit, or introduce an incorrect intermediate state. Such mistakes can then affect later observations and decisions, causing the agent to move further away from the correct solution path. Therefore, while aggressive FP4 quantization is well suited for accelerating the compute-intensive prefill phase, applying it to the whole inference process can destabilize agentic generation.

These observations motivate our phase-aware framework: apply FP4 quantization to the compute-intensive prefill phase to gain efficiency, while retaining high-precision decoding for stable autoregressive generation.

\subsection{Mix-Quant: Quantized Prefilling, Precise Decoding}

\textbf{NVFP4 Microscaling Quantization for Prefilling.}
We adopt NVFP4 weight-and-activation quantization as our quantization method. NVFP4 is a 4-bit microscaling floating-point format~\citep{nvidia2025nvfp4} introduced for Blackwell-generation low-precision tensor-core execution. Each numerical value is represented by an E2M1 FP4 value, while groups of consecutive elements share a local scale. Unlike coarser microscaling formats such as MXFP4, which typically use larger groups and power-of-two scales, NVFP4 adopts smaller groups of 16 elements with FP8 E4M3 block scales, together with an additional tensor-level scale that controls the global dynamic range~\citep{egiazarian2025bridginggappromiseperformance}. This two-level scaling is crucial: the tensor-level scale prevents global saturation, while the local block scale adapts to fine-grained variations within the tensor. Moreover, due to its small group size and fine-grained scaling design, NVFP4 already achieves strong quantization performance with simple round-to-nearest (RTN) quantization, while more complex quantization techniques such as rotation provide little additional benefit and introduce extra runtime overhead~\citep{egiazarian2025bridginggappromiseperformance}. Therefore, we directly adopt RTN quantization in our implementation.

Let $x \in \mathbb{R}^{n}$ be a vectorized activation or weight tensor, and let $\mathcal{B}$ be a partition of its elements into blocks of size $g=16$. For an element $x_i$ in block $b(i)$, NVFP4 quantization can be written as
\begin{equation}
    q_i = \Pi_{\mathrm{FP4}}
    \left(
        \frac{x_i}{\alpha_x \, \sigma_{b(i)}}
    \right),
    \qquad
    \hat{x}_i = \alpha_x \, \sigma_{b(i)} q_i ,
    \label{eq:nvfp4_quant}
\end{equation}
where $\Pi_{\mathrm{FP4}}(\cdot)$ projects a scaled value to the nearest representable FP4 value with clipping, $\sigma_b$ is the FP8 block scale for block $b$, and $\alpha_x$ is the tensor-level scale. A standard amax-based choice sets the block scale according to the largest magnitude in the block,
\begin{equation}
    \sigma_b =
    \Pi_{\mathrm{E4M3}}
    \left(
        \frac{\max_{i \in b} |x_i|}
        {\alpha_x \, q_{\max}}
    \right),
    \label{eq:block_scale}
\end{equation}
where $q_{\max}$ is the largest finite FP4 magnitude and $\Pi_{\mathrm{E4M3}}(\cdot)$ rounds the scale to the FP8 E4M3 grid. In practice, activations and weights use layouts aligned with the GEMM dimension so that quantization, dequantization, and matrix multiplication can be fused efficiently by the backend.

\paragraph{Prefill-Decode Disaggregation Deployment.}
At inference time, Mix-Quant maintains two execution paths for the same base model: an NVFP4 W4A4 prefill path and the original high-precision decode path. We deploy these two paths using a prefill-decode disaggregation framework, where prefill workers process the input prompt and transfer the resulting KV cache to decode workers through a NIXL-based KV-cache transfer mechanism~\citep{kwon2023efficient}. Given a prompt, the quantized prefill path processes all input tokens and writes the initial KV cache in the precision expected by the high-precision decode engine. The decode path then consumes this cache and generates output tokens autoregressively, while KV entries for newly generated tokens are produced by the high-precision decode path as usual, as shown in~\cref{fig:medthod}. This system-level design avoids the extra low-level kernel switching, conversion overhead, and KV-cache misalignment that can arise in mixed-precision quantization pipelines, while preserving the deployment benefits of prefill-decode disaggregation~\citep{zhong2024distserve,patel2023splitwise}.

%% file: sections/experiments.tex
\section{Experiments}
\label{sec:experiments}

\subsection{Experiment Setup}

\paragraph{Benchmarks.}
We evaluate Mix-Quant on a diverse suite of input-intensive benchmarks covering long-context reasoning and agentic inference. For long-context evaluation, we use LongBench-V2~\citep{bai2024longbenchv2} and AA-LCR~\citep{artificialanalysis2025lcr}, which test understanding, synthesis, and reasoning over long documents. For agentic evaluation, we consider BFCL v4~\citep{patil2025berkeley} for tool use and function calling, LongMemEval~\citep{wu2025longmemeval} for long-term interactive memory, $\tau^2$-bench~\citep{barres2025tau} for stateful interactive conversations as general agents. 
To provide a more comprehensive assessment, we further evaluate Mix-Quant on challenging reasoning benchmarks, including Math500~\citep{lightman2023let}, AIME24 and AIME25~\citep{dekoninck2026matharena}.

\paragraph{Models.}
We evaluate recent strong open-weight models for agentic workloads, spanning multiple model families and scales: Qwen3-8B~\citep{yang2025qwen3}, Gemma-4-26B-A4B-it and Gemma-4-31B-it~\citep{gemma4_model_card}, and Qwen3.5-9B~\citep{qwen3.5}. 
These models are selected because they support long-context and agent-oriented use cases and cover both compact and larger-capacity deployment regimes. 
For each model, we compare three variants: the original BF16 model, a uniform NVFP4 W4A4 quantized model, and Mix-Quant.

\paragraph{Evaluation Setup.}
We serve models on RTX 5090 and B200 GPUs to use Blackwell-generation FP4 hardware acceleration. The serving stack is based on vLLM \citep{kwon2023vllm}. For disaggregated execution, we use NIXL-based KV-cache transfer between prefill and decode workers, following the standard disaggregated-prefill serving pattern in which compute-intensive prefill and memory-bandwidth-intensive decode can be placed on separate workers \citep{nvidia2026nixl}. Each benchmark is run independently three times and we report the mean score. We use the default long-context setting of each model family: 256K tokens for Gemma-4 models and 262K tokens for Qwen3.5 models. Since Qwen3-8B natively supports 32K context, we apply YaRN~\citep{peng2024yarn} with scaling factor 4 to extend its context window to 131K tokens for long-context agentic workflows.

\subsection{Main Results}
\begin{table*}[t]
\centering
\small
\setlength{\tabcolsep}{5pt}
\renewcommand{\arraystretch}{1.15}

\caption{Agentic benchmark performance of Mix-Quant across general, long-term memory, and stateful interaction benchmarks. Results are averaged over three independent runs. Best and second-best results within each model group are shown in bold and underlined, respectively.}
\label{tab:agentic}

\resizebox{0.8\textwidth}{!}{
\begin{tabular}{l ccc c}
\toprule
\textbf{Model} & \textbf{BFCL v4} & \textbf{LongMemEval} & \textbf{$\tau^2$-bench} & \textbf{Avg.} \\
\midrule
Qwen3-8B & \underline{40.50} & \textbf{57.00} & \textbf{31.06} & \textbf{42.85}  \\
Qwen3-8B-NVFP4 & 38.77 & 49.82 & 27.34 & 38.64 \\
\textbf{Mix-Quant} & \textbf{40.63} & \underline{54.85} & \underline{28.86} &  \underline{41.45} \\

\midrule
Qwen3.5-9B &
 \textbf{58.96} & \textbf{86.27} & \textbf{86.69} & \textbf{77.31} \\
Qwen3.5-9B-NVFP4 &
 56.86 & 78.00 & 76.26 &  70.37 \\
\textbf{Mix-Quant} &
 \underline{57.89} & \underline{84.27} & \underline{81.89} & \underline{74.68}  \\

\midrule
Gemma-4-26B-A4B-it &
 \textbf{53.07} & \textbf{80.50} & \textbf{64.63} & \textbf{66.07} \\
Gemma-4-26B-A4B-it-NVFP4 &
 48.13 & 62.42 & 57.31 & 55.95 \\
\textbf{Mix-Quant} &
 \underline{51.94} & \underline{72.45} & \underline{60.62} & \underline{61.67} \\

\midrule
Gemma-4-31B-it &
 \underline{68.60} & \textbf{90.80} & \textbf{73.50} & \textbf{77.63} \\
Gemma-4-31B-it-NVFP4 &
 \textbf{68.69} & 89.20 & 70.74 & 76.21 \\
\textbf{Mix-Quant} &
 68.19 & \underline{90.40} & \underline{72.84} & \underline{77.14} \\

\bottomrule
\end{tabular}
}
\end{table*}

\paragraph{Long-Context Agentic Benchmark Results.}
Table~\ref{tab:agentic} reports the results on long-context agentic benchmarks. 
Uniform NVFP4 quantization consistently degrades agentic performance across model families, with average scores dropping from 42.85 to 38.64 for Qwen3-8B, from 77.31 to 70.37 for Qwen3.5-9B, and from 66.07 to 55.95 for Gemma-4-26B-A4B-it. 
These results indicate that directly applying low-precision quantization to the entire inference process can substantially harm long-context agentic reasoning and decision making. 
In contrast, Mix-Quant recovers a large portion of the lost performance by preserving high-precision decoding, achieving average scores of 41.45, 74.68, and 61.67 on the corresponding models, and nearly matching the BF16 baseline on Gemma-4-31B-it with an average score of 77.14 versus 77.63. 
For example, on LongMemEval, uniform NVFP4 causes substantial quality drops for Qwen3-8B and Gemma-4-26B-A4B-it, whereas Mix-Quant improves the scores from 49.82 to 54.85 and from 62.42 to 72.45, respectively. 

This trend is consistent with the phase-aware motivation of Mix-Quant: long-memory and tool-use workloads require efficient processing of large input contexts, while their autoregressive decisions remain highly sensitive to quantization-induced perturbations during decoding.

\paragraph{Reasoning and Long-Context Benchmark Results.}
To complement the agentic evaluation, we further evaluate Mix-Quant on reasoning and long-context benchmarks. These benchmarks assess whether the benefits of Mix-Quant generalize beyond agentic workloads to tasks involving multi-step reasoning and long-context understanding. As shown in Table~\ref{tab:reasoning_longcontext}, Mix-Quant consistently recovers a large portion of the accuracy loss introduced by uniform NVFP4 quantization across these tasks. 
For example, the average score of Qwen3.5-9B drops from 72.04 to 63.26 under uniform NVFP4, while Mix-Quant recovers it to 70.59; for Gemma-4-26B-A4B-it, Mix-Quant nearly matches the BF16 baseline, achieving 71.93 compared with 71.94 under BF16 and 66.31 under uniform NVFP4.
This shows that phase-aware quantization is not only beneficial for agentic serving, but also effective for workloads involving multi-step reasoning and long-context understanding.
This further supports our phase-aware design: aggressive NVFP4 W4A4 quantization is effective for the compute-intensive prefill phase, which primarily encodes the input context and builds the KV cache, while applying the same low-bit policy throughout autoregressive generation process can perturb token decisions and degrade reasoning or long-context generation quality.

\begin{table*}[t]
\centering
\small
\setlength{\tabcolsep}{5pt}
\renewcommand{\arraystretch}{1.15}

\caption{Reasoning and long-context performance of Mix-Quant on mathematical reasoning and long-context benchmarks. Results are averaged over three independent runs.}
\label{tab:reasoning_longcontext}

\resizebox{\textwidth}{!}{
\begin{tabular}{l ccc cc c}
\toprule
\multirow{2}{*}{\textbf{Model}} &
\multicolumn{3}{c}{\textbf{Reasoning}} &
\multicolumn{2}{c}{\textbf{Long Context}} &
\multirow{2}{*}{\textbf{Avg.}} \\
\cmidrule(lr){2-4}\cmidrule(lr){5-6}
& MATH500 & AIME24 & AIME25 & LongBench-V2 & AA-LCR & \\
\midrule

Qwen3-8B & 93.73 & \underline{75.54} & \textbf{67.77} & \textbf{39.86} & \textbf{33.67} & \textbf{62.11} \\
Qwen3-8B-NVFP4 & \underline{94.12} & 66.53 & 55.33 & 35.46 & 24.67 & 55.22 \\
\textbf{Mix-Quant} & \textbf{94.40} & \textbf{76.66} & \underline{66.66} & \underline{38.60} & \underline{28.67} & \underline{61.00} \\

\midrule
Qwen3.5-9B & \textbf{94.85} & \underline{68.89} & \textbf{60.00} & \textbf{55.47} & \textbf{81.00} &  \textbf{72.04} \\
Qwen3.5-9B-NVFP4 & 93.46 & 54.44 & 40.00 & 50.40 & 78.00 & 63.26  \\
\textbf{Mix-Quant} & \underline{94.32} & \textbf{70.33} & \underline{56.67} & \underline{52.29} & \underline{79.33} & \underline{70.59} \\

\midrule
Gemma-4-26B-A4B-it & \textbf{95.86} & \underline{77.67} & \underline{65.33} & \textbf{53.83} & \textbf{67.00} & \textbf{71.94} \\
Gemma-4-26B-A4B-it-NVFP4 & 95.20 & 75.33 & 62.22 & 48.15 &  50.67 & 66.31 \\
\textbf{Mix-Quant} & \underline{95.40} & \textbf{78.67} & \textbf{69.67} & \underline{51.57} & \underline{64.33}  &  \underline{71.93} \\

\midrule
Gemma-4-31B-it & \underline{97.33} & \underline{92.22} & \textbf{82.22} & \textbf{63.35} & \textbf{76.67} & \textbf{82.36} \\
Gemma-4-31B-it-NVFP4 & \textbf{97.75} & 83.33 & 80.00 & 58.55 & 71.67 & 78.26 \\
\textbf{Mix-Quant} & 97.20 & \textbf{93.33} & \underline{81.11} & \underline{61.64} & \underline{73.67} & \underline{81.39} \\

\bottomrule
\end{tabular}
}
\end{table*}

\paragraph{Prefilling Stage Speedup.}

\begin{figure}[t]
    \centering
    \includegraphics[width=1\linewidth]{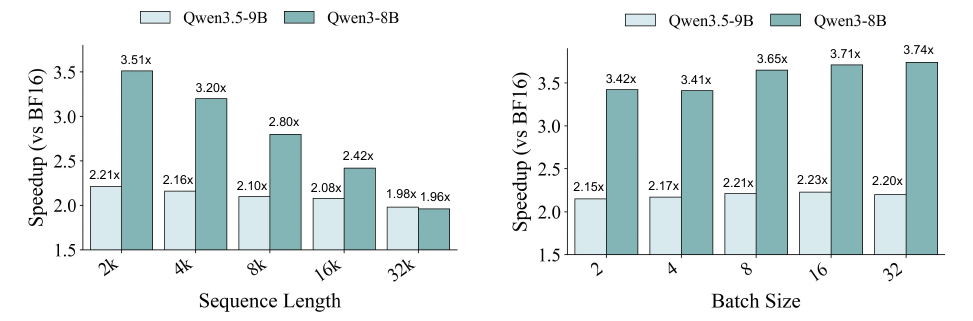}
    \caption{End-to-end prefill latency speedup of Mix-Quant over the BF16 baseline on NVIDIA RTX 5090 GPUs. Left: speedup across different sequence lengths with batch size fixed to 1. Right: speedup across different batch sizes with sequence length fixed to 2K.}
    \label{fig:prefill_speedup}
\end{figure}

We evaluate the end-to-end prefill latency speedup of Mix-Quant over the BF16 baseline on NVIDIA RTX 5090 GPUs. All latency measurements are performed in vLLM, using FlashInfer for attention computation and Blackwell NVFP4 W4A4 GEMM kernels for linear layers. For fair comparison, we keep the batch size, prompt length, KV-cache dtype, and backend configuration identical across methods.

As shown in~\cref{fig:prefill_speedup}, Mix-Quant consistently accelerates prefill across both Qwen3.5-9B and Qwen3-8B under varying sequence lengths and batch sizes, achieving nearly $3\times$ average speedup over BF16. The gains are especially pronounced for Qwen3-8B, while remaining stable for Qwen3.5-9B. These results show that aggressive NVFP4 weight-and-activation quantization can substantially reduce the computation cost of input-intensive prefill workloads, highlighting Mix-Quant's potential for efficient agentic LLM serving.

\paragraph{Phase-wise Quantization Ablation. }

\begin{table*}[t]
\centering
\small
\setlength{\tabcolsep}{5pt}
\renewcommand{\arraystretch}{1.15}

\caption{Phase-wise quantization ablation comparing the impact of quantizing different inference stages on long-context and multi-turn agentic benchmarks. }
\label{tab:phase_ablation}

\resizebox{\textwidth}{!}{
\begin{tabular}{l ccc cc c}
\toprule
\multirow{2}{*}{\textbf{Model}} &
\multicolumn{2}{c}{\textbf{Long Context}} &
\multicolumn{3}{c}{\textbf{Agentic}} &
\multirow{2}{*}{\textbf{Avg.}} \\
\cmidrule(lr){2-3}\cmidrule(lr){4-6}
& LongBench-V2 & AA-LCR & BFCL v4 & LongMemEval & $\tau^2$-bench \\
\midrule

Qwen3-8B & \textbf{39.86} & \textbf{33.67} & \underline{40.50} & \textbf{57.00} & \textbf{31.06} &  \textbf{40.42} \\
Qwen3-8B-NVFP4 & 35.46 & 24.67 & 30.70 & 49.82 & 27.34 &  33.59  \\
Qwen3-8B-P16D4 & 37.38 & \underline{29.33} & 36.40 & 52.47 & 28.13 & 36.74 \\
\textbf{Mix-Quant} & \underline{38.60} & 28.67 & \textbf{40.63} & \underline{54.85} & \underline{28.86} & \underline{38.32} \\

\midrule
Gemma-4-26B-A4B-it &
 \textbf{53.83} & \textbf{67.00} & \textbf{53.07} & \textbf{80.50} & \textbf{64.63} & \textbf{63.81} \\
Gemma-4-26B-A4B-it-NVFP4 &
48.15 & 50.67 & 48.13 & 62.42 & 57.31 & 53.34 \\
Gemma-4-26B-A4B-it-P16D4 & 
50.72 & 63.33 & \underline{52.32} & 71.62 & \underline{61.27} & 59.85 \\
\textbf{Mix-Quant} &
\underline{51.57} & \underline{64.33} & 51.94 & \underline{72.45} & 60.62 & \underline{60.18} \\

\bottomrule
\end{tabular}
}
\end{table*}

In~\cref{tab:phase_ablation}, we compare different phase-wise quantization strategies to isolate the effect of quantizing prefilling and decoding. Mix-Quant applies FP4 quantization only to prefilling while keeping decoding in BF16, whereas P16D4 keeps prefilling in BF16 but performs decoding in FP4. Uniform NVFP4 quantizes both stages.

The results show that prefill-only quantization substantially mitigates the degradation caused by uniform NVFP4. For Qwen3-8B, the average score drops from 40.42 to 33.59 under uniform NVFP4, while Mix-Quant recovers it to 38.32. For Gemma-4-26B-A4B-it, uniform NVFP4 reduces the average score from 63.81 to 53.34, whereas Mix-Quant improves it to 60.18. Compared with decode-only quantization, Mix-Quant also performs better, although the margin is more moderate: it improves the average score over P16D4 from 36.74 to 38.32 on Qwen3-8B, and from 59.85 to 60.18 on Gemma-4-26B-A4B-it. These results suggest that both phase-wise strategies are less harmful than uniform quantization, but quantizing prefilling is generally preferable to quantizing decoding. The advantage is not uniformly large across all benchmarks, since prefill quantization can still perturb hidden states and the KV cache. Nevertheless, decoding remains more sensitive because token-level errors can propagate through subsequent autoregressive steps.

Overall, these results support the phase-aware design of Mix-Quant: apply aggressive FP4 computation to the compute-intensive prefill stage, while preserving higher precision during decoding for fast and stable generation.

%% file: sections/conclusion.tex
\section{Conclusion}
\label{sec:conclusion}

In this work, we identified a critical efficiency-performance dilemma in long-context agentic LLM inference: agentic workflows require repeated processing of long input contexts, making the compute-intensive prefilling phase a major bottleneck, while naively applying low-bit quantization to the full inference pipeline can degrade performance due to error accumulation during autoregressive decoding. To address this challenge, we proposed Mix-Quant, a phase-aware quantization framework that applies high-throughput NVFP4 weight-and-activation quantization to the prefilling phase while preserving BF16 precision for decoding. By decoupling prefilling acceleration from decoding precision, Mix-Quant aligns quantization with the distinct computational characteristics and error sensitivities of different inference stages, thereby improving efficiency without sacrificing generation stability. Extensive experiments on long-context and agentic benchmarks show that Mix-Quant largely preserves task performance while achieving significant throughput gains. These findings demonstrate the promise of phase-aware algorithm–hardware co-design for building efficient and reliable LLM agents.

%% file: neurips_2026.bbl
\begin{thebibliography}{48}
\providecommand{\natexlab}[1]{#1}
\providecommand{\url}[1]{\texttt{#1}}
\expandafter\ifx\csname urlstyle\endcsname\relax
  \providecommand{\doi}[1]{doi: #1}\else
  \providecommand{\doi}{doi: \begingroup \urlstyle{rm}\Url}\fi

\bibitem[{Artificial Analysis Team}(2025)]{artificialanalysis2025lcr}
{Artificial Analysis Team}.
\newblock {Artificial Analysis Long Context Reasoning Benchmark (LCR)}.
\newblock \url{https://artificialanalysis.ai/}, 2025.
\newblock Accessed: 2026-05-06.

\bibitem[Bai et~al.(2024)Bai, Tu, Zhang, Peng, Wang, Lv, Cao, Xu, Hou, Dong, Tang, and Li]{bai2024longbenchv2}
Yushi Bai, Shangqing Tu, Jiajie Zhang, Hao Peng, Xiaozhi Wang, Xin Lv, Shulin Cao, Jiazheng Xu, Lei Hou, Yuxiao Dong, Jie Tang, and Juanzi Li.
\newblock Longbench v2: Towards deeper understanding and reasoning on realistic long-context multitasks.
\newblock \emph{arXiv preprint arXiv:2412.15204}, 2024.

\bibitem[Barres et~al.(2025)Barres, Dong, Ray, Si, and Narasimhan]{barres2025tau}
Victor Barres, Honghua Dong, Soham Ray, Xujie Si, and Karthik Narasimhan.
\newblock {$\tau^2$-Bench}: Evaluating conversational agents in a dual-control environment.
\newblock \emph{arXiv preprint arXiv:2506.07982}, 2025.

\bibitem[Dekoninck et~al.(2026)Dekoninck, Jovanović, Gehrunger, Rögnvalddson, Petrov, Sun, and Vechev]{dekoninck2026matharena}
Jasper Dekoninck, Nikola Jovanović, Tim Gehrunger, Kári Rögnvalddson, Ivo Petrov, Chenhao Sun, and Martin Vechev.
\newblock Beyond benchmarks: Matharena as an evaluation platform for mathematics with llms.
\newblock 2026.
\newblock URL \url{https://arxiv.org/abs/2605.00674}.

\bibitem[Dettmers et~al.(2022)Dettmers, Lewis, Belkada, and Zettlemoyer]{dettmers2022llmint8}
Tim Dettmers, Mike Lewis, Younes Belkada, and Luke Zettlemoyer.
\newblock {LLM.int8()}: 8-bit matrix multiplication for transformers at scale.
\newblock In \emph{Advances in Neural Information Processing Systems}, 2022.

\bibitem[Egiazarian et~al.(2025)Egiazarian, Castro, Kuznedelev, Panferov, Kurtic, Pandit, Marques, Kurtz, Ashkboos, Hoefler, and Alistarh]{egiazarian2025bridginggappromiseperformance}
Vage Egiazarian, Roberto~L. Castro, Denis Kuznedelev, Andrei Panferov, Eldar Kurtic, Shubhra Pandit, Alexandre Marques, Mark Kurtz, Saleh Ashkboos, Torsten Hoefler, and Dan Alistarh.
\newblock Bridging the gap between promise and performance for microscaling fp4 quantization, 2025.
\newblock URL \url{https://arxiv.org/abs/2509.23202}.

\bibitem[Fan et~al.(2026)Fan, Huang, Wu, Wang, Wang, and He]{fan2026flashprefill}
Qihang Fan, Huaibo Huang, Zhiying Wu, Juqiu Wang, Bingning Wang, and Ran He.
\newblock Flashprefill: Instantaneous pattern discovery and thresholding for ultra-fast long-context prefilling.
\newblock \emph{arXiv preprint arXiv:2603.06199}, 2026.
\newblock URL \url{https://arxiv.org/abs/2603.06199}.

\bibitem[Frantar et~al.(2022{\natexlab{a}})Frantar, Ashkboos, Hoefler, and Alistarh]{frantar2022gptq}
Elias Frantar, Saleh Ashkboos, Torsten Hoefler, and Dan Alistarh.
\newblock Gptq: Accurate post-training quantization for generative pre-trained transformers.
\newblock \emph{arXiv preprint arXiv:2210.17323}, 2022{\natexlab{a}}.

\bibitem[Frantar et~al.(2022{\natexlab{b}})Frantar, Ashkboos, Hoefler, and Alistarh]{frantar2023gptq}
Elias Frantar, Saleh Ashkboos, Torsten Hoefler, and Dan Alistarh.
\newblock Gptq: Accurate post-training quantization for generative pre-trained transformers.
\newblock \emph{arXiv preprint arXiv:2210.17323}, 2022{\natexlab{b}}.

\bibitem[{Google DeepMind}(2026)]{gemma4_model_card}
{Google DeepMind}.
\newblock Gemma 4 model card.
\newblock \url{https://ai.google.dev/gemma/docs/core/model_card_4}, 2026.
\newblock Accessed: 2026-05-06.

\bibitem[Gu et~al.(2024)Gu, Dong, Wei, and Huang]{gu2024minillm}
Yuxian Gu, Li~Dong, Furu Wei, and Minlie Huang.
\newblock Minillm: Knowledge distillation of large language models.
\newblock In \emph{The twelfth international conference on learning representations}, 2024.

\bibitem[Jiang et~al.(2024{\natexlab{a}})Jiang, Li, Zhang, Wu, Luo, Ahn, Han, Abdi, Li, Lin, et~al.]{jiang2024minference}
Huiqiang Jiang, Yucheng Li, Chengruidong Zhang, Qianhui Wu, Xufang Luo, Surin Ahn, Zhenhua Han, Amir~H Abdi, Dongsheng Li, Chin-Yew Lin, et~al.
\newblock Minference 1.0: Accelerating pre-filling for long-context llms via dynamic sparse attention.
\newblock \emph{Advances in Neural Information Processing Systems}, 37:\penalty0 52481--52515, 2024{\natexlab{a}}.

\bibitem[Jiang et~al.(2024{\natexlab{b}})Jiang, Wu, Luo, Li, Lin, Yang, and Qiu]{jiang2024longllmlingua}
Huiqiang Jiang, Qianhui Wu, Xufang Luo, Dongsheng Li, Chin-Yew Lin, Yuqing Yang, and Lili Qiu.
\newblock Longllmlingua: Accelerating and enhancing llms in long context scenarios via prompt compression.
\newblock In \emph{Proceedings of the 62nd Annual Meeting of the Association for Computational Linguistics (Volume 1: Long Papers)}, pages 1658--1677, 2024{\natexlab{b}}.

\bibitem[Jimenez et~al.(2023)Jimenez, Yang, Wettig, Yao, Pei, Press, and Narasimhan]{jimenez2023swe}
Carlos~E Jimenez, John Yang, Alexander Wettig, Shunyu Yao, Kexin Pei, Ofir Press, and Karthik Narasimhan.
\newblock Swe-bench: Can language models resolve real-world github issues?
\newblock \emph{arXiv preprint arXiv:2310.06770}, 2023.

\bibitem[Kwon et~al.(2023{\natexlab{a}})Kwon, Li, Zhuang, Sheng, Zheng, Yu, Gonzalez, Zhang, and Stoica]{kwon2023efficient}
Woosuk Kwon, Zhuohan Li, Siyuan Zhuang, Ying Sheng, Lianmin Zheng, Cody~Hao Yu, Joseph~E. Gonzalez, Hao Zhang, and Ion Stoica.
\newblock Efficient memory management for large language model serving with pagedattention.
\newblock In \emph{Proceedings of the ACM SIGOPS 29th Symposium on Operating Systems Principles}, 2023{\natexlab{a}}.

\bibitem[Kwon et~al.(2023{\natexlab{b}})Kwon, Li, Zhuang, Sheng, Zheng, Yu, Gonzalez, Zhang, and Stoica]{kwon2023vllm}
Woosuk Kwon, Zhuohan Li, Siyuan Zhuang, Ying Sheng, Lianmin Zheng, Cody~Hao Yu, Joseph~E. Gonzalez, Hao Zhang, and Ion Stoica.
\newblock Efficient memory management for large language model serving with pagedattention.
\newblock In \emph{Proceedings of the ACM SIGOPS 29th Symposium on Operating Systems Principles}, 2023{\natexlab{b}}.

\bibitem[Li et~al.(2025)Li, Su, Yang, Xie, Wang, Xie, Wong, and Yang]{li2025quantization}
Zhen Li, Yupeng Su, Runming Yang, Congkai Xie, Zheng Wang, Zhongwei Xie, Ngai Wong, and Hongxia Yang.
\newblock Quantization meets reasoning: Exploring llm low-bit quantization degradation for mathematical reasoning.
\newblock \emph{arXiv preprint arXiv:2501.03035}, 2025.

\bibitem[Lightman et~al.(2023)Lightman, Kosaraju, Burda, Edwards, Baker, Lee, Leike, Schulman, Sutskever, and Cobbe]{lightman2023let}
Hunter Lightman, Vineet Kosaraju, Yuri Burda, Harrison Edwards, Bowen Baker, Teddy Lee, Jan Leike, John Schulman, Ilya Sutskever, and Karl Cobbe.
\newblock Let's verify step by step.
\newblock In \emph{The Twelfth International Conference on Learning Representations}, 2023.

\bibitem[Lin et~al.(2024)Lin, Tang, Tang, Yang, Chen, Wang, Xiao, Dang, Gan, and Han]{lin2024awq}
Ji~Lin, Jiaming Tang, Haotian Tang, Shang Yang, Wei-Ming Chen, Wei-Chen Wang, Guangxuan Xiao, Xingyu Dang, Chuang Gan, and Song Han.
\newblock Awq: Activation-aware weight quantization for on-device llm compression and acceleration.
\newblock \emph{Proceedings of machine learning and systems}, 6:\penalty0 87--100, 2024.

\bibitem[Lu et~al.(2025)Lu, Fang, Ma, Li, and Wang]{lu2025mixreasoning}
Haiquan Lu, Gongfan Fang, Xinyin Ma, Qi~Li, and Xinchao Wang.
\newblock Mixreasoning: Switching modes to think.
\newblock \emph{arXiv preprint arXiv:2510.06052}, 2025.

\bibitem[Luo et~al.(2025)Luo, Zhang, Yuan, Zhao, Yang, Gu, Wu, Chen, Qiao, Long, et~al.]{luo2025large}
Junyu Luo, Weizhi Zhang, Ye~Yuan, Yusheng Zhao, Junwei Yang, Yiyang Gu, Bohan Wu, Binqi Chen, Ziyue Qiao, Qingqing Long, et~al.
\newblock Large language model agent: A survey on methodology, applications and challenges.
\newblock \emph{arXiv preprint arXiv:2503.21460}, 2025.

\bibitem[Ma et~al.(2023)Ma, Fang, and Wang]{ma2023llm}
Xinyin Ma, Gongfan Fang, and Xinchao Wang.
\newblock Llm-pruner: On the structural pruning of large language models.
\newblock \emph{Advances in neural information processing systems}, 36:\penalty0 21702--21720, 2023.

\bibitem[Nakano et~al.(2021)Nakano, Hilton, Balaji, Wu, Ouyang, Kim, Hesse, Jain, Kosaraju, Saunders, Jiang, Cobbe, Eloundou, Krueger, Button, Knight, Chess, and Schulman]{nakano2021webgpt}
Reiichiro Nakano, Jacob Hilton, Suchir Balaji, Jeff Wu, Long Ouyang, Christina Kim, Christopher Hesse, Shantanu Jain, Vineet Kosaraju, William Saunders, Xu~Jiang, Karl Cobbe, Tyna Eloundou, Gretchen Krueger, Kevin Button, Matthew Knight, Benjamin Chess, and John Schulman.
\newblock Webgpt: Browser-assisted question-answering with human feedback.
\newblock \emph{arXiv preprint arXiv:2112.09332}, 2021.

\bibitem[{NVIDIA}(2025)]{nvidia2025nvfp4}
{NVIDIA}.
\newblock Introducing nvfp4 for efficient and accurate low-precision inference.
\newblock \url{https://developer.nvidia.com/blog/introducing-nvfp4-for-efficient-and-accurate-low-precision-inference/}, 2025.

\bibitem[{NVIDIA}(2026)]{nvidia2026nixl}
{NVIDIA}.
\newblock Enhancing distributed inference performance with the nvidia inference transfer library, 2026.
\newblock Accessed 2026-05-03.

\bibitem[Packer et~al.(2023)Packer, Wooders, Lin, Fang, Patil, Stoica, and Gonzalez]{packer2023memgpt}
Charles Packer, Sarah Wooders, Kevin Lin, Vivian Fang, Shishir~G. Patil, Ion Stoica, and Joseph~E. Gonzalez.
\newblock Memgpt: Towards llms as operating systems.
\newblock \emph{arXiv preprint arXiv:2310.08560}, 2023.

\bibitem[Patel et~al.(2023)Patel, Choukse, Zhang, Shah, Goiri, Maleki, and Bianchini]{patel2023splitwise}
Pratyush Patel, Esha Choukse, Chaojie Zhang, Aashaka Shah, Inigo Goiri, Saeed Maleki, and Ricardo Bianchini.
\newblock Splitwise: Efficient generative llm inference using phase splitting.
\newblock \emph{arXiv preprint arXiv:2311.18677}, 2023.

\bibitem[Patel et~al.(2024)Patel, Choukse, Zhang, Shah, Goiri, Maleki, and Bianchini]{patel2024splitwise}
Pratyush Patel, Esha Choukse, Chaojie Zhang, Aashaka Shah, {\'I}{\~n}igo Goiri, Saeed Maleki, and Ricardo Bianchini.
\newblock Splitwise: Efficient generative llm inference using phase splitting.
\newblock In \emph{2024 ACM/IEEE 51st Annual International Symposium on Computer Architecture (ISCA)}, pages 118--132. IEEE, 2024.

\bibitem[Patil et~al.(2025)Patil, Mao, Yan, Ji, Suresh, Stoica, and Gonzalez]{patil2025berkeley}
Shishir~G Patil, Huanzhi Mao, Fanjia Yan, Charlie Cheng-Jie Ji, Vishnu Suresh, Ion Stoica, and Joseph~E Gonzalez.
\newblock The berkeley function calling leaderboard (bfcl): From tool use to agentic evaluation of large language models.
\newblock In \emph{Forty-second International Conference on Machine Learning}, 2025.

\bibitem[Peng et~al.(2024)Peng, Quesnelle, Fan, and Shippole]{peng2024yarn}
Bowen Peng, Jeffrey Quesnelle, Honglu Fan, and Enrico Shippole.
\newblock Yarn: Efficient context window extension of large language models.
\newblock In \emph{International Conference on Learning Representations}, 2024.

\bibitem[Qiao et~al.(2025)Qiao, Yao, Rajbhandari, and He]{qiao2025swiftkv}
Aurick Qiao, Zhewei Yao, Samyam Rajbhandari, and Yuxiong He.
\newblock Swiftkv: Fast prefill-optimized inference with knowledge-preserving model transformation.
\newblock In \emph{Proceedings of the 2025 Conference on Empirical Methods in Natural Language Processing}, pages 25745--25764, 2025.

\bibitem[{Qwen Team}(2026)]{qwen3.5}
{Qwen Team}.
\newblock {Qwen3.5}: Towards native multimodal agents, February 2026.
\newblock URL \url{https://qwen.ai/blog?id=qwen3.5}.

\bibitem[Schick et~al.(2023)Schick, Dwivedi-Yu, Dessi, Raileanu, Lomeli, Zettlemoyer, Cancedda, and Scialom]{schick2023toolformer}
Timo Schick, Jane Dwivedi-Yu, Roberto Dessi, Roberta Raileanu, Maria Lomeli, Luke Zettlemoyer, Nicola Cancedda, and Thomas Scialom.
\newblock Toolformer: Language models can teach themselves to use tools.
\newblock \emph{arXiv preprint arXiv:2302.04761}, 2023.

\bibitem[Wadlom et~al.(2026)Wadlom, Shen, and Lu]{wadlom2026efficient}
Noppanat Wadlom, Junyi Shen, and Yao Lu.
\newblock Efficient llm serving for agentic workflows: A data systems perspective.
\newblock \emph{arXiv preprint arXiv:2603.16104}, 2026.

\bibitem[Wu et~al.(2025{\natexlab{a}})Wu, Wang, Yu, Zhang, Chang, and Yu]{wu2025longmemeval}
Di~Wu, Hongwei Wang, Wenhao Yu, Yuwei Zhang, Kai-Wei Chang, and Dong Yu.
\newblock Longmemeval: Benchmarking chat assistants on long-term interactive memory.
\newblock In \emph{International Conference on Learning Representations}, 2025{\natexlab{a}}.

\bibitem[Wu et~al.(2025{\natexlab{b}})Wu, Xiao, Nie, Guo, Lou, Wong, Mo, Zhang, Forys, Ai, et~al.]{wu2026plena}
Haoran Wu, Can Xiao, Jiayi Nie, Xuan Guo, Binglei Lou, Jeffrey~TH Wong, Zhiwen Mo, Cheng Zhang, Przemyslaw Forys, Chengyang Ai, et~al.
\newblock Combating the memory walls: Optimization pathways for long-context agentic llm inference.
\newblock \emph{arXiv preprint arXiv:2509.09505}, 2025{\natexlab{b}}.

\bibitem[Xiao et~al.(2023)Xiao, Lin, Seznec, Wu, Demouth, and Han]{xiao2023smoothquant}
Guangxuan Xiao, Ji~Lin, Mickael Seznec, Hao Wu, Julien Demouth, and Song Han.
\newblock Smoothquant: Accurate and efficient post-training quantization for large language models.
\newblock In \emph{International Conference on Machine Learning}, 2023.

\bibitem[Xu et~al.(2025)Xu, Liang, Mei, Gao, Tan, and Zhang]{xu2025mem}
Wujiang Xu, Zujie Liang, Kai Mei, Hang Gao, Juntao Tan, and Yongfeng Zhang.
\newblock A-mem: Agentic memory for llm agents.
\newblock \emph{arXiv preprint arXiv:2502.12110}, 2025.

\bibitem[Yang et~al.(2025{\natexlab{a}})Yang, Li, Yang, Zhang, Hui, Zheng, Yu, Gao, Huang, Lv, Zheng, Liu, Zhou, Huang, Hu, Ge, Wei, Lin, Tang, Yang, Tu, Zhang, Yang, Yang, Zhou, Zhou, Lin, Dang, Bao, Yang, Yu, Deng, Li, Xue, Li, Zhang, Wang, Zhu, Men, Gao, Liu, Luo, Li, Tang, Yin, Ren, Wang, Zhang, Ren, Fan, Su, Zhang, Zhang, Wan, Liu, Wang, Cui, Zhang, Zhou, and Qiu]{qwen3}
An~Yang, Anfeng Li, Baosong Yang, Beichen Zhang, Binyuan Hui, Bo~Zheng, Bowen Yu, Chang Gao, Chengen Huang, Chenxu Lv, Chujie Zheng, Dayiheng Liu, Fan Zhou, Fei Huang, Feng Hu, Hao Ge, Haoran Wei, Huan Lin, Jialong Tang, Jian Yang, Jianhong Tu, Jianwei Zhang, Jianxin Yang, Jiaxi Yang, Jing Zhou, Jingren Zhou, Junyang Lin, Kai Dang, Keqin Bao, Kexin Yang, Le~Yu, Lianghao Deng, Mei Li, Mingfeng Xue, Mingze Li, Pei Zhang, Peng Wang, Qin Zhu, Rui Men, Ruize Gao, Shixuan Liu, Shuang Luo, Tianhao Li, Tianyi Tang, Wenbiao Yin, Xingzhang Ren, Xinyu Wang, Xinyu Zhang, Xuancheng Ren, Yang Fan, Yang Su, Yichang Zhang, Yinger Zhang, Yu~Wan, Yuqiong Liu, Zekun Wang, Zeyu Cui, Zhenru Zhang, Zhipeng Zhou, and Zihan Qiu.
\newblock Qwen3 technical report.
\newblock \emph{arXiv preprint arXiv:2505.09388}, 2025{\natexlab{a}}.

\bibitem[Yang et~al.(2025{\natexlab{b}})Yang, Li, Yang, Zhang, Hui, Zheng, Yu, Gao, Huang, Lv, et~al.]{yang2025qwen3}
An~Yang, Anfeng Li, Baosong Yang, Beichen Zhang, Binyuan Hui, Bo~Zheng, Bowen Yu, Chang Gao, Chengen Huang, Chenxu Lv, et~al.
\newblock Qwen3 technical report.
\newblock \emph{arXiv preprint arXiv:2505.09388}, 2025{\natexlab{b}}.

\bibitem[Yang et~al.(2024{\natexlab{a}})Yang, Jimenez, Wettig, Lieret, Yao, Narasimhan, and Press]{yang2024swe}
John Yang, Carlos~E Jimenez, Alexander Wettig, Kilian Lieret, Shunyu Yao, Karthik Narasimhan, and Ofir Press.
\newblock Swe-agent: Agent-computer interfaces enable automated software engineering.
\newblock \emph{Advances in Neural Information Processing Systems}, 37:\penalty0 50528--50652, 2024{\natexlab{a}}.

\bibitem[Yang et~al.(2024{\natexlab{b}})Yang, Jimenez, Wettig, Lieret, Yao, Narasimhan, and Press]{yang2024sweagent}
John Yang, Carlos~E. Jimenez, Alexander Wettig, Kilian Lieret, Shunyu Yao, Karthik Narasimhan, and Ofir Press.
\newblock Swe-agent: Agent-computer interfaces enable automated software engineering.
\newblock \emph{arXiv preprint arXiv:2405.15793}, 2024{\natexlab{b}}.

\bibitem[Yao et~al.(2022)Yao, Zhao, Yu, Du, Shafran, Narasimhan, and Cao]{yao2022react}
Shunyu Yao, Jeffrey Zhao, Dian Yu, Nan Du, Izhak Shafran, Karthik Narasimhan, and Yuan Cao.
\newblock React: Synergizing reasoning and acting in language models.
\newblock \emph{arXiv preprint arXiv:2210.03629}, 2022.

\bibitem[Yao et~al.(2023)Yao, Zhao, Yu, Du, Shafran, Narasimhan, and Cao]{yao2023react}
Shunyu Yao, Jeffrey Zhao, Dian Yu, Nan Du, Izhak Shafran, Karthik Narasimhan, and Yuan Cao.
\newblock React: Synergizing reasoning and acting in language models.
\newblock In \emph{International Conference on Learning Representations}, 2023.

\bibitem[Zhao et~al.(2025)Zhao, Lu, Wang, Kong, and Wu]{zhao2025qspec}
Juntao Zhao, Wenhao Lu, Sheng Wang, Lingpeng Kong, and Chuan Wu.
\newblock Qspec: Speculative decoding with complementary quantization schemes.
\newblock In \emph{Proceedings of the 2025 Conference on Empirical Methods in Natural Language Processing}, pages 4779--4795, 2025.

\bibitem[Zhao et~al.(2024)Zhao, Lin, Zhu, Ye, Chen, Zheng, Ceze, Krishnamurthy, Chen, and Kasikci]{zhao2024atom}
Yilong Zhao, Chien-Yu Lin, Kan Zhu, Zihao Ye, Lequn Chen, Size Zheng, Luis Ceze, Arvind Krishnamurthy, Tianqi Chen, and Baris Kasikci.
\newblock Atom: Low-bit quantization for efficient and accurate llm serving.
\newblock \emph{arXiv preprint arXiv:2310.19102}, 2024.

\bibitem[Zhong et~al.(2024)Zhong, Liu, Chen, Hu, Zhu, Liu, Jin, and Zhang]{zhong2024distserve}
Yinmin Zhong, Shengyu Liu, Junda Chen, Jianbo Hu, Yibo Zhu, Xuanzhe Liu, Xin Jin, and Hao Zhang.
\newblock Distserve: Disaggregating prefill and decoding for goodput-optimized large language model serving.
\newblock In \emph{18th USENIX Symposium on Operating Systems Design and Implementation}, 2024.

\bibitem[Zhou et~al.(2024)Zhou, Ning, Hong, Fu, Xu, Li, Lou, Wang, Yuan, Li, et~al.]{zhou2024survey}
Zixuan Zhou, Xuefei Ning, Ke~Hong, Tianyu Fu, Jiaming Xu, Shiyao Li, Yuming Lou, Luning Wang, Zhihang Yuan, Xiuhong Li, et~al.
\newblock A survey on efficient inference for large language models.
\newblock \emph{arXiv preprint arXiv:2404.14294}, 2024.

\end{thebibliography}
